\pdfoutput=1

\documentclass[11pt]{article}

\usepackage{acl}

\usepackage{times}
\usepackage{latexsym}

\usepackage[T1]{fontenc}

\usepackage[utf8]{inputenc}

\usepackage{microtype}

\usepackage{graphicx}
\usepackage{subfigure}
\usepackage{amsmath}
\usepackage{booktabs}
\usepackage{multirow}
\usepackage{colortbl}
\usepackage{bm}

\title{Graph-to-Text Generation with Dynamic Structure Pruning}

\author{Liang Li$^{1, 2}$, Ruiying Geng$^{3}$, Bowen Li$^{3}$, Can Ma${}^1$\thanks{$^{*}$Corresponding authors: Can Ma, Yongbin Li}, Yinliang Yue$^{1}$, \\ \textbf{Binhua Li}$^{3}$ \textbf{,} \textbf{Yongbin Li}$^{3*}$ \\
$^1$Institute of Information Engineering, Chinese Academy of Sciences, Beijing, China \\
$^2$School of Cyber Security, University of Chinese Academy of Sciences, Beijing, China\\
$^3$DAMO Academy, Alibaba Group \\
\texttt{\{liliang, macan, yueyinliang\}@iie.ac.cn} \\
\texttt{\{ruiying.gry, binhua.lbh, shuide.lyb\}@alibaba-inc.com}
}
\begin{document}
\maketitle
\begin{abstract}
Most graph-to-text works are built on the encoder-decoder framework with cross-attention mechanism. 
Recent studies have shown that explicitly modeling the input graph structure can significantly improve the performance. 
However, the vanilla structural encoder cannot capture all specialized information in a single forward pass for all decoding steps, resulting in inaccurate semantic representations. 
Meanwhile, the input graph is flatted as an unordered sequence in the cross attention, ignoring the original graph structure. 
As a result, the obtained input graph context vector in the decoder may be flawed. 
To address these issues, we propose a Structure-Aware Cross-Attention (SACA) mechanism to re-encode the input graph representation conditioning on the newly generated context at each decoding step in a structure aware manner. We further adapt SACA and introduce its variant Dynamic Graph Pruning (DGP) mechanism to dynamically drop irrelevant nodes in the decoding process.
We achieve new state-of-the-art results on two graph-to-text datasets, LDC2020T02 and ENT-DESC, with only minor increase on computational cost.

\end{abstract}

\section{Introduction}

\begin{figure}[t]
\centering
\includegraphics[width=1.0\columnwidth]{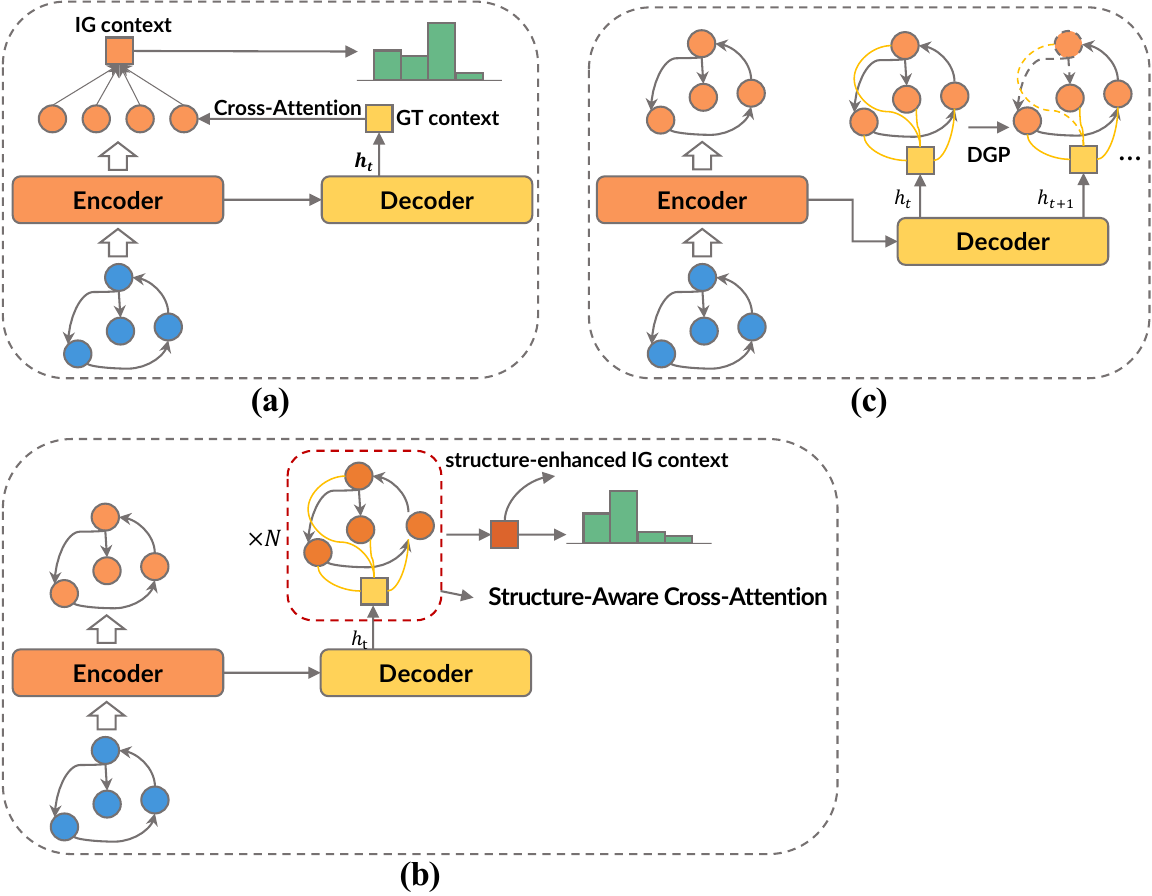}
\caption{(a) denotes an encoder-decoder framework with the cross-attention mechanism where IG and GT contexts denote the input graph and generated text graph contexts, respectively. (b) is an example of Structure-Aware Cross-Attention. The dotted lines in (c) denote the pruned edges and nodes.}
\label{fig1:introduction}
\end{figure}

Data-to-text task aims to generate a natural language description from structural or semi-structural data, such as tables \cite{DBLP:conf/emnlp/WisemanSR17}, Abstract Meaning Representation (AMR) graphs \cite{banarescu2013abstract}, and Knowledge Graphs (KG) \cite{cheng-etal-2020-ent}. It helps people get the key points of the input data and makes the stored information accessible to a broader audience of end-users. There have been several practical application scenarios
in this field, such as biography generation \cite{lebret2016neural}, basketball news generation \cite{DBLP:conf/emnlp/WisemanSR17}, and advertising text generation \cite{shao2019long}.
This paper focuses on generation from graph structures in AMR and KG, referred to as graph-to-text.

In recent years, encoder-decoder with the cross-attention mechanism has been the de facto framework for graph-to-text tasks (shown in Figure \ref{fig1:introduction}(a)). 
Given an input graph, the encoder first computes vector representations for the graph nodes. 
On the decoding side, Input Graph (IG) context vector is obtained via cross-attention based on the partially Generated Text (GT) at each time step, then the next target token is finally predicted. 
Unlike conventional text-to-text tasks, the structural nature of the input graph makes it unsuitable to naively apply sequential encoder-decoder architecture to the graph-to-text task. 
To alleviate this issue, recent studies \cite{song2018graph,damonte2019structural,cai2020graph} proposed to utilize the graph encoder to capture the input graph structure. These works have demonstrated that explicitly modeling the graph structure can bring benefits to the model performance.

Although equipped with the structure-aware modeling, it is still hard for the encoder to capture all specialized information for graph-to-text generation.
It is evidenced by recent studies \citep{liu2019hierarchical, li-etal-2021-improving-encoder} that a vanilla structural encoder cannot capture the accurate semantic representation of the input structural data effectively.
Auxiliary supervision has been shown to be helpful, but effective auxiliary tasks are not easy to design and may not generalize well to different datasets.
We suspect that it is challenging for the encoder to encode all relevant information into node representations in a single forward pass for all the decoding steps, especially if the input graph structure is complex. Besides the encoder side, few works have focused on the decoder side for graph-to-text tasks.
Considering the ordinary cross-attention mechanism, the representations of input data obtained from the encoder are still treated as an \emph{unordered} node representation sequence.
We conjuncture that this \emph{plain} cross-attention does not take full advantage of the input graph structure and therefore may harm the model performance.

Current models with graph encoder and cross-attention may yield inaccurate input graph context representation due to the deficiency on both encoder and decoder as we discussed before.
To tackle the above problems and avoid introducing auxiliary tasks, we propose a novel \textbf{Structure-Aware Cross-Attention} (SACA) mechanism. 
Apart from the \emph{plain} cross-attention, our SACA re-encodes the input graph conditioning on the newly generated context in a \emph{structure-aware} fashion. 
Other than a single forward pass, specialized representations from the source side are built adaptively at each decoding step, which makes the decoder easily exploit relevant-only information for prediction.
More specifically, as shown in Figure \ref{fig1:introduction}(b), we construct a joint graph, in which we explicitly treat the generated text context vector as an additional node and connect it with nodes in the input graph at each decoding step.
We implement SACA using the relational graph attention network (RGAT, \citealt{shaw2018self}). 
Furthermore, we stack multiple layers of SACAs to perform deep interactions between the generated text context vector and input node representations.
Finally, we fetch the node representation corresponding to the newly added node as the structure-enhanced input graph context to predict the target token. 

In practice, we notice that some nodes become irrelevant and uninformative as the decoding goes on. 
These nodes are distracting and can disturb the generation process. 
Intuitively, the decoder should dynamically discard the unrelated parts of the graph at different decoding steps. In other words, the joint graph structure should be dynamically adjusted. 
To this end, we adapt SACA and propose its variant \textbf{Dynamic Graph Pruning} (DGP) mechanism (shown in Figure \ref{fig1:introduction}(c)). 
DGP prunes the structure of the joint graph via the gate mechanism to achieve sparse connections between the nodes based on the generated text context.

We conduct experiments on two graph-to-text datasets, LDC2020T02\footnote{https://catalog.ldc.upenn.edu/LDC2020T02} and ENT-DESC \cite{cheng-etal-2020-ent}, to verify the effectiveness of the proposed approach. 
Empirical results show that our proposed methods achieve new state-of-the-art results on the two datasets. 
Further experiments indicate that SACA and DGP do not reduce the diversity of the generated text and can better handle complex graphs. 
Meanwhile, additional investigation reveals that SACA and DGP only bring minor increase on the model size and inference time.

\section{Related Works}
Graph-to-text is a challenging task which aims at generating a descriptive text from the structured knowledge, such Knowledge Graph (KG), and Abstract Meaning Representation (AMR) graphs. It is helpful for interpretability of KGs in general \cite{schmitt-etal-2020-unsupervised} and knowledge-based question answering \cite{hui-etal-2022-s2sql, wang2022proton,fu2020survey,qin2022survey}.

In recent years, most graph-to-text methods have been built based on the encoder-decoder architecture. This kind of method usually consists of a structural encoder and a decoder. The structural encoder aims to model the structure information into the representation of the input graph. \citet{song2018graph} first propose the graph recurrent networks (GRNs) to encode the AMR node directly. And then, some works \cite{shi2020g2t,chenkgpt} introduce the Graph Neural Networks (GNNs) as the structural encoder, which updates the representations of nodes based on their immediate neighbors. To integrate both local and non-local features and learn a better structural representation of a graph, \citet{guo2019densely} introduce the dense connection, allowing deeper GCNs. 
Unlike the local information aggregation scheme, \citet{zhu2019modeling,cai2020graph} propose the Graph Transformer that uses explicit relation encoding and allows direct communication between two distant nodes.

A recently proposed neural abstractive Multi-Document Summarization (MDS) model, GraphSumm \cite{DBLP:conf/acl/LiXLWWD20}, also considers the input graph structure during decoding. The biggest difference between Graphsum and our proposed SACA is that the former only introduces one graph attention layer in each decoder layer. SACA, on the other hand, injects graph structure into decoding by re-encoding the input graph. Specifically, it re-computes the input graph representation by conditioning it on the newly generated text at each decoding step.

Recent approaches try to apply the Pre-trained Language Models (PLMs) \cite{kenton2019bert,raffel2019exploring} into the graph-to-text generation. 
Particularly, \citet{ribeiro-etal-2021-structural} propose to utilize the adapter method \cite{pfeiffer-etal-2020-mad} to encode graph structure into PLMs and only train graph structure-aware adapter parameters. In this way, they avoid catastrophic forgetting while maintaining the topological structure of the graph.

\begin{figure*}[t]
\centering
\includegraphics[width=2\columnwidth]{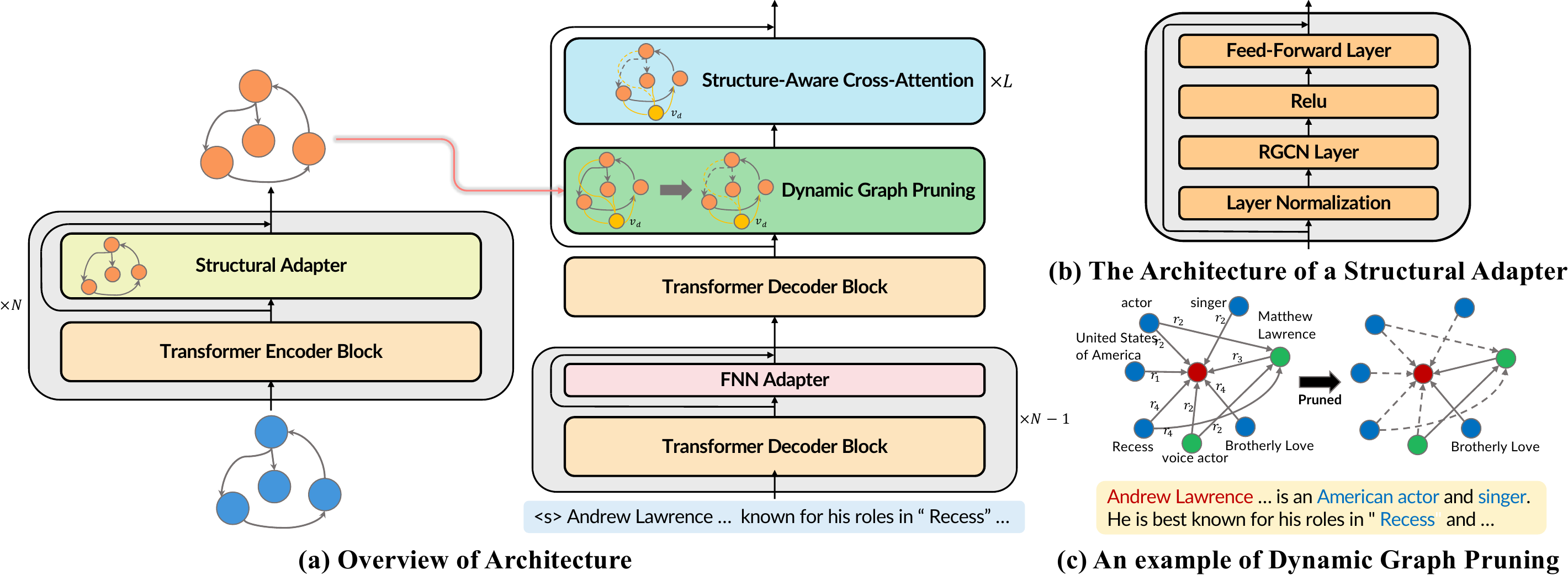}
\caption{Illustration of the proposed model architecture. (a) is an overview of our model. (b) is the architecture of a structural adapter. (c) is an example of Dynamic Graph Pruning, where $r_1 \sim r_4$ denote the relations: ``\textbf{country of citizenship}", ``\textbf{occupation}'', ``\textbf{sibling}'', and ``\textbf{cast member}'', respectively. The dummy lines in (c) denote the pruned edges.}
\label{fig:overall}
\end{figure*}

\section{Approach}
\label{section_model}
We expect that developing graph-to-text generation should benefit from the recent advance on pre-trained language models (PLMs) \cite{lewisbart,raffel2019exploring}.
To explicitly encode the input graph structure into PLMs while alleviating the catastrophic forgetting problem, we consider SA-RGCN \cite{ribeiro-etal-2021-structural} as our baseline model. SA-RGCN is an adapter method to encode graph structure into PLMs. 
The overall illustration of our model architecture is shown in Figure \ref{fig:overall}(a). 
In this section, we first introduce how to represent the input graph and the architecture of our baseline SA-RGCN. 
Then, we depict our proposed Structure-Aware Cross-Attention (SACA) in details.
Lastly, we adapt SACA and propose its variant Dynamic Graph Pruning (DGP).

\subsection{Graph Representation}
\label{section_graph}
Let $\mathcal{G}_0 = (\mathcal{V}_0, \mathcal{E}_0, \mathcal{R}_0)$ denote a multi-relational and directed graph with nodes $v_i \in  \mathcal{V}_0$ and labeled edges $(v_i,r,v_j) \in \mathcal{E}_0$, where $r \in \mathcal{R}_0$ is the relation type. Following previous work \cite{beck2018graph}, we convert each input graph into a Levi graph $\mathcal{G}_l = (\mathcal{V}_l, \mathcal{E}_l)$, which is an unlabeled and connected bipartite graph. Specifically, each labeled edge $(v_i , r, v_j ) \in \mathcal{E}_0$ is transformed into two unlabeled edges $(v_i, r), (r, v_j ) \in \mathcal{E}_l$. In addition, we add a reverse edge $(v_j, v_i)$ for each default edge $(v_i, v_j)$. Therefore, each Levi graph $\mathcal{G}_l$ contains two type relations $\mathcal{R}_l = \{d, r\}$, where $d$ and $r$ denote the default and reverse edge, respectively. To better take advantage of the PLMs, we convert each $\mathcal{G}_l$ into a new token graph $\mathcal{G} = (\mathcal{V}, \mathcal{E}, \mathcal{R})$, where each token of a node in $\mathcal{V}_l$ becomes a node $v \in \mathcal{V}$. 
\subsection{Pretrained LMs with Structural Adapters}
\label{section_sa}
To inject graph structural bias into PLMs, we incorporate the structural adapter \cite{ribeiro-etal-2021-structural} into the PLMs encoder. As shown in Figure \ref{fig:overall} (a), we add a structural adapter after each transformer encoder block on the encoder. Figure \ref{fig:overall} (b) illustrates the architecture of a structural adapter, in where a relational GCN (RGCN) \cite{DBLP:conf/esws/SchlichtkrullKB18} layer computes the node representation based on the local neighborhood of node $v \in \mathcal{V}$. Formally, at each layer $l$, given the encoder layer representation $h^l_v$, a structural adapter computes the representation for $v$ by the following:
\begin{align}
g^l_v &= \sum_{r \in \mathcal{R}} \sum_{u \in N_r(v)}
  \frac
    {1}
    {|\mathcal{N}_r(v)|}
  W^l_r \textrm{LN}(h^l_v),
\\
z^l_v &= W^l_e (\sigma(g^l_v)) + h^l_v,
\end{align}
where $\textrm{LN}(\cdot)$ denotes layer normalization. $\mathcal{N}_r(v)$ is the sef of immediate neighbors under relation $r \in \mathcal{R}$. $W^l_r$ encodes the edge type between the nodes $u$ and $v$. $\sigma$ is the activation function.

We add an FNN adapter after each transformer decoder block to adapt the language model to the graph-to-text task. Given the output $\hat{h}^l_v$ of the $l$ th transformer decoder block, the adapter representation is computed as:
\begin{align}
\hat{z}^l_v &= W^l_o ({\sigma(W^l_p \textrm{LN}( \hat{h}^l_v ))}) + \hat{h}^l_v,
\end{align}
where $W^l_o$ and $W^l_d$ denote learnable parameters.

\subsection{Structure-Aware Cross-Attention}
\label{section_saca}
We argue that the input graph context representation obtained by the \emph{plain} cross-attention may be inaccurate. 
The reason is twofold.
First, it is not easy for the graph encoder to capture all specialized information required for generation in a single forward pass. Therefore, a single encoder without any auxiliary assistant may not be effective in capturing the accurate semantic representation \cite{liu2019hierarchical, li-etal-2021-improving-encoder}. In other words, the graph representation encoded by the graph encoder may be inaccurate. Second, during decoding, the decoder treats structural data as an unordered node sequence, which ignores the input graph structure. However, the graph structure has been proven to play an essential role in the graph representation and may offer clues about which nodes are more related to the generated text context.

To tackle the above challenge, we propose a Structure-Aware Cross-Attention (SACA) mechanism, which re-encodes the input graph representation by conditioning on the newly generated context. Specifically, we first build a joint graph, in which we view the generated text (GT) context as a new node $v_d$ and explicitly connect it to each node in the input graph $\mathcal{G}$ at each decoding step. The corresponding reverse edges are also added. The joint graph can be formulated as $\mathcal{G}_{joint} = (\mathcal{V}_{joint}, \mathcal{E}_{joint}, \mathcal{R}\textbf{})$, where $\mathcal{V}_{joint} = \{v_d\} \cup \mathcal{V}$ and $\mathcal{E}_{joint} = \{ (v_i, v_d), (v_d, v_i); v_i \in \mathcal{V} \} \cup \mathcal{E}$. We use the representations from the encoder for the node from $\mathcal{V}$ and the hidden state from the last transformer decoder block as the representation for the GT context node.

To induce the representations for the nodes in the joint graph $\mathcal{G}_{joint}$ and facilitate introducing Dynamic Graph Pruning (in Section 3.4), we consider graph neural network built on graph attention framework (GAT) \cite{shaw2018self}. Moreover, we employ the relational graph attention network (RGAT) implemented by \citet{shaw2018self} to model the relation between neighbor nodes. Specifically, at each RGAT layer $l$, we update the representation $h^l_v$ of each node $v \in \mathcal{G}_{joint}$ by:
\begin{align}
s_{v, u} &= \frac
  {
    W^q {h^l_v}^{T}
    (W^k h^l_u + E^r_{\mathcal{R}(v,u)})}
  {\sqrt{m}},
  \\
\alpha_{v, u} &= \frac{e^{s_{v, u}}}{\sum_{u' \in \mathcal{N}_{v}} e^{s_{v, u'}}},  \label{eq:softmax}
\\
h^{l+1}_v &=  \sigma(\sum_{u \in \mathcal{N}_v} \alpha_{v, u} W_v h^l_u ),
\end{align}
where $E^r_{\mathcal{R}(v,u)}$ means the embedding of the relation between node $v$ and $u$. $m$ denotes the hidden dimension of RGAT. Finally, the representation vector $h^L_{v_d}$ corresponding to the GT context node $v_d$ is fetched and used as the structure-enhanced input graph context vector for token prediction. 

In conclusion, SACA provides two advantages. First, it re-encodes the input graph by conditioning its representation on the newly generated context. As a result, we build specialized representations which make it easier for the decoder to exploit relevant-only information for prediction at each decoding step. Second, the re-encoding explicitly injects structural bias into input graph context modeling, helping the decoder obtain a more accurate input graph context vector. 
The proposed SACA can be plugged after the last transformer decoder block as shown in Figure \ref{fig:overall} (a).

\subsection{Dynamic Graph Pruning}
\label{section_dgp}
In practice, we notice that some nodes become irrelevant and uninformative as the decoding goes on. These unrelated nodes are distracting and can even disturb the subsequent generation. Intuitively, the decoder should dynamically prune the joint graph at different decoding steps. For this purpose, we adapt SACA and propose its variant Dynamic Graph Pruning (DGP) mechanism, which aims to dynamically drop the redundant nodes in the joint graph according to the generated text during decoding. The DGP employs the gate mechanism to sparse the connection between a node and its immediate neighbors in the joint graph to achieve graph pruning. Specifically, at each decoding step $t$, for each node $v$ in the joint graph, we formulate its gate as bellow:
\begin{align}
\label{equation_dgp}
{g}_v &= sigmoid(W^T_g tanh(W_e h_v + W_d h^d_t)),
\end{align}
where $W_g$, $W_e$, and $W_d$ are learnable parameters. And $h_v$ is the representation of node $v$ and $h^d_t$ is the decoder hidden state at decoding step $t$, which is usually considered as the representation of the generated text context. The value of gate $g_v \in R$ decides whether the node $v_i$ should be dropped or not. 
Correspondingly, we apply the gate value to multiple SACA layers invariably by modifying the attention weights in SACA (Equation~\ref{eq:softmax}) as follows:
\begin{align}
\alpha_{v, u} &= \frac{g_{u} \odot e^{s_{v, u}}}{\sum_{u' \in \mathcal{N}_{v}} g_{u'} \odot e^{s_{v, u'}}}.
\end{align}

Intuitively, if the value of gate $g_u$ is close to $0$, the connections between node $u$ with all its immediate neighbors will be largely weaken. That is, the node is removed from the joint graph. 
Specifically, the attention score $\alpha_{v, u}$ measures the relevance between any two nodes, $v$ and $u$, in the joint graph, while the gate $g_v$ models the relevance between the node $v$ and the generated text context $h_t$.

As a shown example in Figure \ref{fig:overall} (c), the red node represents the main entity. Initially, the main entity connects with all its neighbor nodes. As the decoding goes on, some nodes are redundant for the subsequent generation. For example, the nodes ``actor`` has been described, and node ``voice actor`` is also covered by the generated text. Therefore, DGP discards these nodes by giving them gates with small values.

We observed that the values of the gates calculated by Equation \ref{equation_dgp} are almost equal to $1$, indicating that the model does not actively learn to prune a graph. Inspired by \citet{xue2020not}, we further introduce a regularization item, encouraging the network to turn off more gates and generate more sparse connections between nodes in the input graph. We formulate it as follows:
\begin{align}
L_{DGP} &= \frac{\sum^{|y|}_{t=1}\Vert Gate_t \Vert_{1}}{|y|},
\end{align}
 where $Gate_t = \{g_v| v \in \mathcal{V} \}$. $\Vert*\Vert_{1}$ means $L1$ norm regularizer.

\begin{table}
\begin{center}
\resizebox{\columnwidth}{!}{
\begin{tabular}{l|c|c}
\hline
 & \bf ENT-DESC & \bf LDC2020T02 \\
 \hline
 \#train/dev/test & 88,650/11,081/11,081 & 55,635/1,722/1,898 \\
 \#relations & 967 & 157 \\
 Avg \#nodes & 18.0 & 14.2 \\
 Avg \#triples & 27.4 & 14.8 \\
 Avg length & 31.0 & 95.0\\
\hline
\end{tabular}
}
\end{center}
\caption{Dataset statistics of ENT-DESC and LDC2020T02.}
\label{tab:datasets}
\end{table} 

\begin{table*}[t]
\small
\begin{center}
\resizebox{\textwidth}{!}{
\begin{tabular}{llllll}
\hline
\rowcolor[RGB]{237,237,237} & \multicolumn{5}{c}{\bf LDC2020T02}\\

\rowcolor[RGB]{237,237,237} \multirow{-2}{*}{\bf Models} & BLEU & METEOR & ChRF++ & $\boldsymbol{\mathcal{M}}$ & BERTScore \\
\hline
LDGCN \cite{zhang2020lightweight} & 34.3 & 38.2 & 63.7 & - & - \\
SPRING \cite{DBLP:conf/aaai/BevilacquaBN21} & 44.9 & - & 72.9 & - & - \\
FINETUNE \cite{ribeiro-etal-2021-structural} & 41.6$_{\pm 0.6}$ & - & 70.4$_{\pm 0.5}$ & 78.5$_{\pm 0.2}$ & 96.0$_{\pm 0.1}$  \\
ADAPT \cite{ribeiro-etal-2021-structural} & 43.0$_{\pm 0.2}$  & - & 71.3$_{\pm 0.1}$& 79.3$_{\pm 0.1}$  & 96.2$_{\pm 0.1}$   \\
SA-RGCN \cite{ribeiro-etal-2021-structural} & 48.0$_{\pm 0.2}$ & - & 73.2$_{\pm 0.1}$ & 80.1$_{\pm 0.3}$  & 96.3$_{\pm 0.1}$  \\
\hline
FINETUNE$^{\ddagger}$ & 41.55$_{\pm 0.58}$  & 42.06$_{\pm 0.21}$ & 70.62$_{\pm 0.34}$ & 78.30$_{\pm 0.32}$  & 96.02$_{\pm 0.12}$\\
\textbf{SA-RGCN}$^{\ddagger}$ & 47.85$_{\pm 0.22}$ & 45.11$_{\pm 0.16}$ & 73.53$_{\pm 0.19}$ & 80.31$_{\pm 0.24}$  & 96.41$_{\pm 0.03}$  \\
\textbf{Ours} & \bf 48.78$_{\pm 0.08}$ & \bf46.12$_{\pm 0.12}$& \bf 74.35$_{\pm 0.09}$ & \bf 80.69$_{\pm 0.41}$  & \bf 96.62$_{\pm 0.02}$   \\
\hline
\rowcolor[RGB]{237,237,237} & \multicolumn{5}{c}{\bf  ENT-DESC} \\
\rowcolor[RGB]{237,237,237} \multirow{-2}{*}{\bf Models} & BLEU & METEOR & ChRF++ & ROUGE-L & PARENT \\
\hline
S2S \cite{bahdanau2015neural} & 6.8 & 10.8 & - & 40.7 & 10.0 \\
GraphTransformer \cite{koncel2019text} & 19.1 & 16.1 & - & 54.3 & 21.4 \\
GRN \cite{beck2018graph} & 24.4 & 18.9 & - & 55.5 & 21.3 \\
GCN \cite{marcheggiani2018deep} & 24.8 & 19.3 & - & 56.2  & 21.8 \\
DeepGCN \cite{guo2019densely} & 24.9 & 19.3 & - & 56.2  & 21.8 \\
MGCN + CNN \cite{cheng-etal-2020-ent} & 26.4 & 20.4 & - & 57.4 & 24.2 \\
\hline
FINETUNE$^{\ddagger}$ & 32.39$_{\pm 0.12}$ & 30.39$_{\pm 0.02}$ & 53.87$_{\pm 0.06}$ & 56.27$_{\pm 0.05}$ & 42.35$_{\pm 0.18}$ \\
\textbf{SA-RGCN}$^{\ddagger}$ & 34.06$_{\pm 0.31}$ & 31.54$_{\pm 0.04}$ & 57.78$_{\pm 0.06}$ & 58.42$_{\pm 0.04}$  & 43.32$_{\pm 0.18}$ \\
\textbf{Ours} & \bf 34.87$_{\pm 0.36}$ & \bf 32.37$_{\pm 0.11}$ & \bf 58.41$_{\pm 0.22}$ & \bf 58.97$_{\pm 0.14}$ & \bf 43.70$_{\pm 0.12}$ \\
\hline
\end{tabular}
}
\end{center}
\caption{Main results of models on LDC2020T02 and ENT-DESC test datasets. $^{\ddagger}$ means our reimplementation. The other results are copied from the original paper. Mean ($\pm$s.d.) over 4 seeds.}
\label{tab:ent_man_results}
\end{table*}

\subsection{Training}
\label{section_training}
Given a reference output $y = \{y_1, y_2, ..., y_T \}$ and an input graph $\mathcal{G}$, we use the cross-entropy loss as the objective function of graph-to-text generation:
\begin{align}
L_{lm} &= \sum^{|y|}_{t = 1} \log p(y_t|y_{1:t-1}, \mathcal{G}, \theta).
\end{align}

Finally, the overall objective function consists of two parts:
\begin{align}
\label{equation:loss}
L &= L_{lm} + \lambda L_{DGP},
\end{align}
where $\lambda$ is a tunable hyper-parameter and is used to make a trade-off between the cross-entropy loss and the regularization item. Intuitively, the $L_{DGP}$ object encourages the model to learn how to prune the graph, and the $L_{lm}$ trains the model to generate the text according to the graph and restrains DGP from pruning too much.

\section{Experiments}
\subsection{Datasets}
We demonstrate the effectiveness of our models on two graph-to-text datasets: LDC2020T02 and ENT-DESC \cite{cheng-etal-2020-ent} 
LDC2020T02 is an AMR-to-Text dataset and has 55,635/1,722/1,898 instances for training, development, and testing. We follow \citet{ribeiro-etal-2021-structural} to preprocess the AMR graphs and tokenize the sentences. Each instance contains a sentence and an AMR graph. ENT-DESC is a large-scale and challenging dataset generating text from the Knowledge Graph (KG-to-Text). Each instance contains a KG consisting of a main entity and a few topic-related entities. The target text consists of sentences that verbalize the main entity in KG. ENT-DESC lacks explicit alignment between the input and the output. Therefore, some knowledge in the input graph may be noise. We follow official training, development, and test splits of 88,650/11,081/11,081 instances. Table \ref{tab:datasets} summarizes the detailed statistics of LDC2020T02 and ENT-DESC.

\subsection{Settings}
Our implementation is based on Hugging Face \cite{wolf2019huggingface}. The RGCN and RGAT are implemented based on PyTorch Geometric \cite{Fey/Lenssen/2019}. We initialize our models by T5 \cite{raffel2019exploring}. To make a fair comparision, we following the same experimental setting with SA-RGCN \cite{ribeiro-etal-2021-structural}. We set the hidden dimensions of both \emph{Structural Adapter} and SACA to $256$. And we use T5$_{base}$ for all experiments on ENT-DESC and T5$_{large}$ on LDC2020T02 for a fair comparison with baselines. We use the AdamW optimizer \cite{loshchilov2018decoupled} and employ a linearly decreasing learning rate schedule without warm-up. The learning rate is fixed as $10^{-4}$. We set the training batch size as $4$ for all experiments. We freeze the T5 parameters and only update the newly added parameters during training. We tune the hyper-parameter $\lambda$ in Equation \ref{equation:loss} from the set $[1^{-2}, 5^{-3}, 1^{-3}, 5^{-4}]$, and select the best one on the development set. We stack $L=2$ RGAT layers in Structure-Aware Cross-Attention. During decoding, we use beam search with a beam size $5$. We use BLEU \cite{papineni2002bleu} for the early stopping criterion. All experiments are trained on Nvidia Tesla V100 32GB GPUs.

Following previous works, on both datasets, we evaluate the results with BLEU \cite{papineni2002bleu}, METEOR \cite{denkowski2011meteor}, and ChRF++ \cite{popovic2015chrf} on both datasets. On LDC2020T02, following  \citet{ribeiro-etal-2021-structural}, we utilize the meaning ($\mathcal{M}$) component of the $\mathcal{MF}$-score \cite{DBLP:conf/eacl/OpitzF21} to measure how well the source AMR graph can be reconstructed from the generated sentence (refer to \ref{append:mf_score} for more details). We use BERTScore \cite{DBLP:conf/iclr/ZhangKWWA20} allowing a semantic evaluation that depends less on the surface forms. On ENT-DESC, We add ROUGE-L \cite{lin2004rouge} and employ PARENT \cite{DBLP:conf/acl/DhingraFPCDC19} for evaluating the faithfulness. We conduct experiments over 4 different seeds and report the average scores on them. 

\subsection{Main Results}
We compare our method with recent state-of-the-art methods (please refer to \ref{baseline} for more details). Table~\ref{tab:ent_man_results} summarizes the results on LDC2020T02 and ENT-DESC test sets. FINETUNE is a method that transforms the input graph into a sequence and finetunes T5 directly. It does not consider the input graph structure. For LDC2020T02, our method outperforms the previous state-of-the-art model by $0.78$ BLEU and $1.15$ ChRF++. Compared with our implemented SA-RGCN, we improve $1.01$ METEOR. Moreover, our method raises $0.38$ $\boldsymbol{\mathcal{M}}$, which indicates that it can generate more faithful sentences to the input graphs. The improvement on \emph{BERTScore} shows that the sentence generated by our method is more similar to the ground truth on the semantic level. For ENT-DESC, we notice FINETUNE performs better than all previous methods. SA-RGCN, which encodes graph structure into T5, furtherly improves the performance. And our model exceeds all previous works and achieves new state-of-the-art results on all metrics. The above results indicate that our proposed methods can improve the model on fluency and faithfulness.

\begin{table}
\begin{center}
\resizebox{\columnwidth}{!}{
\begin{tabular}{lcccccc}
\hline
\rowcolor[RGB]{237,237,237} {\bf Models} & \bf BLEU & \bf METEOR & $\boldsymbol{\mathcal{M}}$ & \bf Dis-1 & \bf Dis-2 \\
\hline
\textbf{GOLD} & - & - & 81.00 & 23.82 & 71.76 \\ 
\hline

\textbf{ADAPT} & 45.22 & 43.28 & 79.56 & 23.20 & 71.40 \\
\hline
\textbf{Ours} & \bf 47.85 & \bf 45.80 & \bf 80.37 & 23.46 & 71.75 \\

\textbf{w/o DGP} &  47.68 & 45.51  & 80.21 & 23.51 & 72.08\\
\textbf{w/o SACA \& DGP} & 47.20 & 45.05 & 80.01 & 23.38 & 71.69 \\

\textbf{w/o StrucAdapt} & 45.43 & 43.54 & 79.75 & 23.32 & 71.65 \\
\hline

\end{tabular}
}
\end{center}
\caption{Ablation study of models on LDC2020T02 development dataset. \textbf{GOLD} indicates the ground-truth sentences. Dis-1 and Dis-2 denote Distinct1 and Distinct2, respectively.}
\label{tab:amr}
\end{table}

\subsection{Analysis and Discussion}
\paragraph{Ablation Study}
The overall performance on the two datasets shows the superiority of our proposed Structure-Aware Cross-Attention (SACA) and Dynamic Graph Pruning (DGP). To demonstrate the effectiveness of each component, we conduct ablation studies on LDC2020T02 development sets and minus one particular component at a time to understand its impact on the performance. Especially, \textbf{w/o DGP} denotes we remove the dynamic graph pruning module and the training objective $L_{DGP}$. \textbf{ADAPT} and \textbf{w/o StrucAdapt} denote replacing each structural adapter in SA-RGCN's and our encoders with an FNN adapter, respectively. \textbf{W/o StrucAdapt} means that the model only considers the structural information during decoding. The results are summarized in Table~\ref{tab:amr}. Particularly, we observe the performance drops after removing SACA or DGP. This indicates that injecting the structural information into input graph context modeling (SACA) and dynamically removing the redundant nodes (DPG) are beneficial.
Regarding the $\mathcal{M}$ score, our model and \textbf{ADAPT} are close to \textbf{GOLD}.
The AMR parser utilized by $\mathcal{M}$, \textbf{ADAPT} as well as our method are all initialized by T5. 
And the AMR paring and AMR-to-Text are dual tasks actually.
Therefore, the $\mathcal{M}$ score is biased and the results of our model and \textbf{ADAPT} are somehow inflated.
Additionally, we utilize Distinct-1 and Distinct-2 \cite{DBLP:conf/naacl/LiGBGD16} to evaluate the diversity of the output text. We observe that SACA and DGP have little effect on Distinct-1 and Distinct-2. This implies that they will not reduce the diversity of the output text.

We notice that, compared with \textbf{ADAPT}, \textbf{w/o StrucAdapt} shows a slight improvement. This indicates it is necessary to explicitly model the graph structure in the encoder, even though structural bias has been injected into the input graph context modeling during decoding. We believe this may be attributed to SACA relying on the input graph representation encoded by the encoder. Because our SACA is designed to exploit the relevant-only information for prediction, it re-encodes the input graph by conditioning its representation on the newly generated context. Therefore, the initial representation for the input graph is important.

\paragraph{Impact on the parameter and speed}
Furthermore, we investigate the impact of SACA and DPG on the model parameters and inference speed on LDC2020T02 development. Specifically, we calculate the additional parameters of each model with respect to T5$_{large}$. And we set the batch size to $1$ to calculate the average decoding time for generating all examples. The results summarized in Table~\ref{tab:param_and_speed} indicate that SACA and DGP only bring minor increase on the model size and inference time.

\begin{table}
\small
\begin{center}
\resizebox{\columnwidth}{!}{
\begin{tabular}{lcc}

\hline
\rowcolor[RGB]{237,237,237} \bf Models & \bf \# Additional Params (million) & \bf Latency (s) \\
\hline
\textbf{ADAPT} & 28.72 (3.3\%) & 1.41 \\
\hline
\textbf{SA-RGCN} & 37.80 (4.9\%) & 1.49 \\
\quad \textbf{+ SACA} & 39.21 (5.0\%) & 1.54 \\ 
\quad \textbf{+ SACA \& DGP} & 41.31 (5.0\%) & 1.55 \\ 
\hline
\end{tabular}
}
\end{center}
\caption{Impact on parameter and speed.}
\label{tab:param_and_speed}
\end{table}

\begin{table}[t!]
    \centering
\resizebox{\columnwidth}{!}{
\begin{tabular}{llll}
\hline
\rowcolor[RGB]{237,237,237} \bf Graph Size & 1-30& 31-60 & $>$60 \\
\rowcolor[RGB]{237,237,237} \bf \# Examples & 840 & 678 & 380 \\
\hline
SA-RGCN & 54.10 & 44.89 & 46.12 \\
Ours &    54.55\bm{$_{+0.45}$} & 45.88\bm{$_{+0.99}$} & 46.72\bm{$_{+0.60}$}  \\
\hline
\rowcolor[RGB]{237,237,237} \bf Graph Diameter & 1-8& 9-12 & $>$12 \\
\rowcolor[RGB]{237,237,237} \bf \# Examples & 824 & 603 & 471 \\
\hline
SA-RGCN & 56.98 & 43.12 & 46.07 \\
Ours & 57.01\bm{$_{+0.03}$} & 43.59\bm{$_{+0.47}$} & 46.99\bm{$_{+0.92}$} \\
\hline
\rowcolor[RGB]{237,237,237} \bf Reentrancies & $<=1$ & 2 & $>$2 \\
\rowcolor[RGB]{237,237,237} \bf \# Examples & 913 & 549 & 436 \\
\hline
SA-RGCN & 53.60 & 44.03 & 43.30 \\
Ours & 54.16\bm{$_{+0.56}$} & 44.55\bm{$_{+0.52}$} & 44.53\bm{$_{+1.23}$} \\
\hline
\end{tabular}
}
    \caption{BLEU scores with respect to graph size, graph diameter and number of reentrancies on LDC2020T02 test set.}
    \label{tab:graph_stru}
\end{table}

\paragraph{Impact on the Graph Properties}
To examine the robustness of our proposed methods, we investigate the model's performance concerning different graph properties (graph size, graph diameter, and reentrancies) on LDC2020T02 and ENT-DESC. Following previous works \cite{cheng-etal-2020-ent,ribeiro-etal-2021-structural}, we use BLEU as the metric. The results are summarized in Table~\ref{tab:graph_stru} and Table~\ref{tab:graph_stru_ent_dest}, respectively. For LDC2020T02, we firstly note that the BLEU scores decrease as the graph size increases since the larger graph is often complex. Our method achieves a clear improvement when handling graphs with $> 30$ nodes. And then we observe that the BLEU gap between our method and SA-RGCN becomes larger for a relatively larger graph diameter. Reentrancies are the nodes with multiple parents. A graph with more reentrancies is typically more complex \cite{wang2020amr}. As shown in the last section in Table \ref{tab:graph_stru}, our method has an improvement of $+1.23$ BLEU points compared to SA-RGCN when graphs contain $>2$ reentrancies. To sum up, the results on the LDC2020T02 dataset show the advantage of our model in dealing with the AMR graph with more complex structures.

As shown in Table \ref{tab:graph_stru_ent_dest}, both models perform differently on ENT-DESC than on LDC2020T02. First, we notice that both models perform the best when the graph size is between $31$ and $50$, and they perform poorly when the graph size is too small or too large. \citet{cheng-etal-2020-ent} also observed the finding, and they believe this is due to the insufficient or very noisy input information for generation. Additionally, both models perform better when graph diameter or number of the reentrancies increase. 
The reason is that, in the ENT-DESC, the knowledge graph with a small diameter or number of the reentrancies contains more noisy information for the generation. Please refer to \ref{append:distribution_on_graph_size} for more details. The BLEU gap between our method and SA-RGCN is the largest when the graph diameter $> 5$ or the number of reentrancies $> 10$. The above results demonstrate that our approach makes SA-RGCN better at handling complex knowledge graphs.

\begin{table}[t!]
    \centering
\resizebox{\columnwidth}{!}{
\begin{tabular}{llll}
\hline
\rowcolor[RGB]{237,237,237} \bf Graph Size & 1-20 & 21-40 & $>$40 \\
\rowcolor[RGB]{237,237,237} \bf \# Examples & 3,559 & 5,069 & 2,453 \\
\hline
SA-RGCN & 33.01 & 38.86 & 28.54 \\
Ours &    33.67 \bm{$_{+0.66}$} & 39.44\bm{$_{+0.58}$} & 29.02 \bm{$_{+0.48}$}  \\
\hline
\rowcolor[RGB]{237,237,237} \bf Graph Diameter & 1-3 & 4-5 & $>$5 \\
\rowcolor[RGB]{237,237,237} \bf \# Examples & 2,227 & 5,017 & 3,787 \\
\hline
SA-RGCN & 30.52 & 34.41 & 35.83 \\
Ours & 31.14\bm{$_{+0.62}$} & 34.83\bm{$_{+0.45}$} & 36.55\bm{$_{+0.72}$} \\
\hline
\rowcolor[RGB]{237,237,237} \bf Reentrancies & $<6$ & 6-10 & $>$10 \\
\rowcolor[RGB]{237,237,237} \bf \# Examples & 2,277 & 5,017 & 3,787 \\
\hline
SA-RGCN & 27.57 & 36.58 & 37.17 \\
Ours & 28.03 \bm{$_{+0.46}$} & 37.17 \bm{$_{+0.59}$} & 37.81\bm{$_{+0.64}$} \\
\hline
\end{tabular}
}
    \caption{BLEU scores with respect to graph size, graph diameter and number of reentrancies on ENT-DEST test set.}
    \label{tab:graph_stru_ent_dest}
\end{table}

We investigate how the model behaves on different types of graphs (AMR and KG). And the results demonstrate that our model deals better with complex structures. We believe the improvement comes from two aspects. First, on the one hand, it is challenging for an encoder to encode all relevant information into node representations in a single forward pass, especially if the graph structure is complex. On the other hand, the re-encoding in SACA makes the decoder easily exploit the relevant-only information for prediction and explicitly injects the structural information at each decoding step. Second, DGP dynamically removes the nodes which are redundant for the subsequent generation, which makes the decoder pay more attention to the relevant nodes.

\begin{table}
\small
\begin{center}
\resizebox{\columnwidth}{!}{
\begin{tabular}{lccc}

\hline
\rowcolor[RGB]{237,237,237} \bf Models & \bf BLEU & \bf METEOR & \bf ROUGE-L \\
\hline
GraphWriter & 14.30 & 18.80 & - \\
\hline
GraphWriter$^{\ddagger}$ & 14.13 {\small $\pm$ 0.10} & 18.92 {\small $\pm$ 0.28} & 27.61 {\small $\pm$ 0.16} \\
Ours & \textbf{15.59} {\small $\pm$0.35} & \textbf{19.70} {\small $\pm$0.21} & \textbf{28.47} {\small $\pm$0.14} \\
\hline
\end{tabular}
}
\end{center}
\caption{Generalization Study on AGENDA test dataset. $^{\ddagger}$ means our reimplementation.}
\label{tab:agenda}
\end{table}

\subsection{Generalization Study}
Institutionally, our proposed methods can not only be applied to PLMs but also RNN based models. In other words, we can easily combine the SACA and DGP with previous RNN based works. To examine the generalization of SACA and DGP, we choose GraphWriter \cite{koncel2019text} as the baseline, which consists of a multi-layer graph transformer encoder and an attention-based decoder with a copy mechanism. Further, to make a fair comparison, we conduct the generalization experiment on AGENDA dataset \cite{koncel2019text}. We simply replace the \emph{plain} cross-attention in GraphWriter with our proposed SACA. Additionally, we add the DGP layer before the SACA. The experiments are under the same settings as described in GraphWriter. As shown in Table \ref{tab:agenda}, we observe that our proposed model significantly improves the performance of GraphWriter. The results indicate that SACA and DGP are not only effective well on PLMs-based models but also potent for RNN-based models. 

\subsection{Human Evaluation}
\begin{figure}[t]
\centering
\includegraphics[width=0.8\columnwidth]{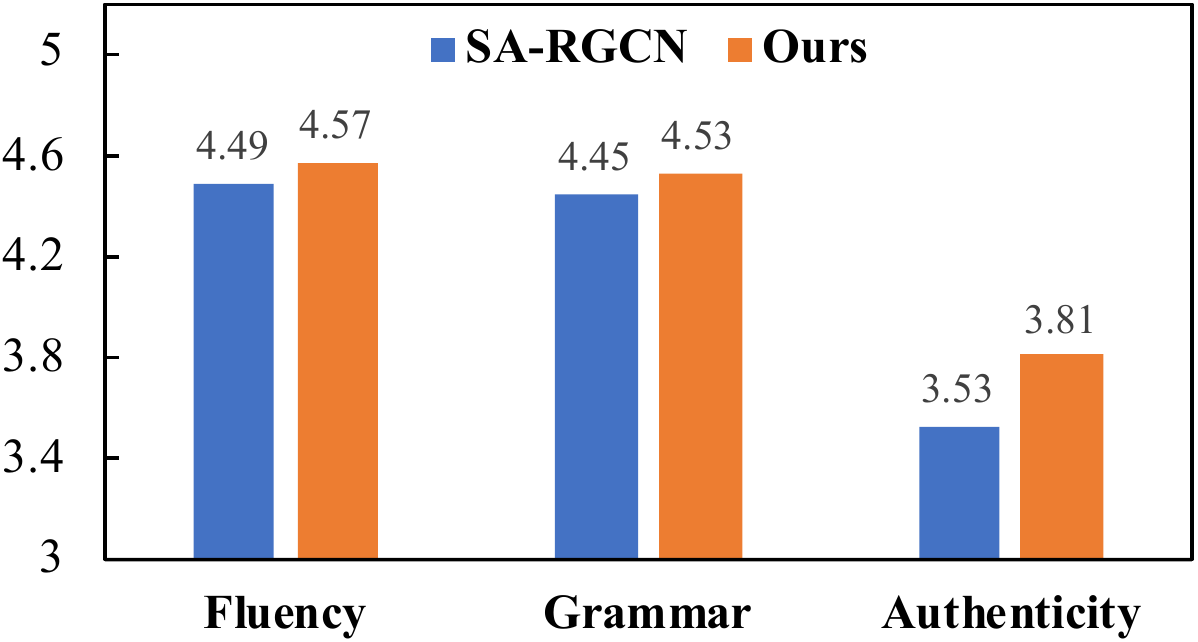}
\caption{Human evaluation results on ENT-DESC test set.}
\label{fig:human_evaluation}
\end{figure}

Considering that the knowledge graph is more readable than AMR, we do human evaluations on the ENT-DESC test set to examine whether human judgments corroborate improvements in automatic evaluation metrics. Following \citet{cheng-etal-2020-ent},   from outputs generated by the baseline model SA-RGCN and our final model (Ours). We distribute the outputs of different systems to three annotators with linguistic backgrounds. The annotators have no knowledge in advance about which model the generated text comes from. Specifically, we give each participant all main entities' neighbors, 1-hop and 2-hop connections between main entities, and topic-related entities as references.
They are required to score the generated text from 1 to 5 in terms of three criteria: Fluency (is the sentence fluent?), Grammar (is the sentence grammatical?), and Authenticity (is the sentence more related to the input graph?). For each criterion, we calculate the final score by averaging the scores from all annotators. As shown in Figure~\ref{fig:human_evaluation}, our model outperforms the baseline SA-RGCN on Fluency and Grammar metrics. For Authenticity, the improvement is more significant. The performance validates the benefit of our proposed SACA and DGP modules in capturing more accurate input graph context representations. We supply a case study in \ref{append:case_study}.
 
\section{Conclusions}
In this work, we make two main contributions. First, we propose Structure-Aware Cross-Attention (SACA) to make decoder easily exploit relevant-only information for prediction. Apart from the plain cross-attention, SACA re-encodes the input graph conditioning on the newly generated context while explicitly considering the input graph structure.
The second one is that we adapt SACA and propose its variant Dynamic Graph Pruning (DGP) mechanism. In detail, the DGP dynamically prunes the structure of the joint graph at different decoding steps according to the generated text. Experimental results conducted on two graph-to-text datasets, LDC2020T02 and ENT-DESC, show the effectiveness of our method. The empirical and analysis results on both datasets show that the proposed methods can improve the model's performance on complex graphs while only bringing minor increase on the model size and inference time.

\bibliography{anthology,custom_v2}
\bibliographystyle{acl_natbib}

\clearpage
\appendix

\section{Appendix}
\label{sec:appendix}
\subsection{$\mathcal{MF}$-score}
\label{append:mf_score}
 The $\mathcal{M}$ (Meaning Preservation) component of the $\mathcal{MF}$-score \cite{DBLP:conf/eacl/OpitzF21} is utilized to measure how well the source AMR graph can be reconstructed from the generated sentence. It reconstructs the AMR with a SOTA parser and computes the relative graph overlap of the reconstruction and the source AMR using graph matching. $\mathcal{M}$ employs the python library amrlib\footnote{https://github.com/bjascob/amrlib/tree/0.5.0} (version $0.5.0$) to make AMR parse, where the parser is a T5-based model.
\begin{figure*}
    \centering
    \subfigure[The distribution of graph  diameter by graph size.]{
    \begin{minipage}[t]{0.4\linewidth}
    \centering
    \includegraphics[width=1\columnwidth]{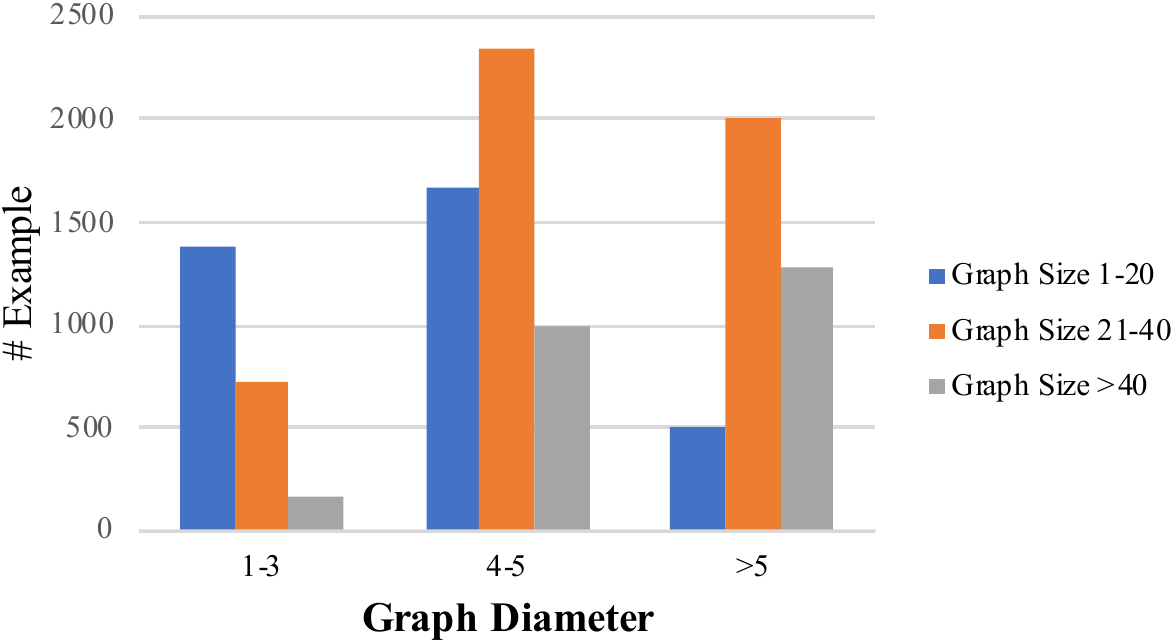}
    \end{minipage}
    
        \begin{minipage}[t]{0.3\linewidth}
    \centering
    \includegraphics[width=1\columnwidth]{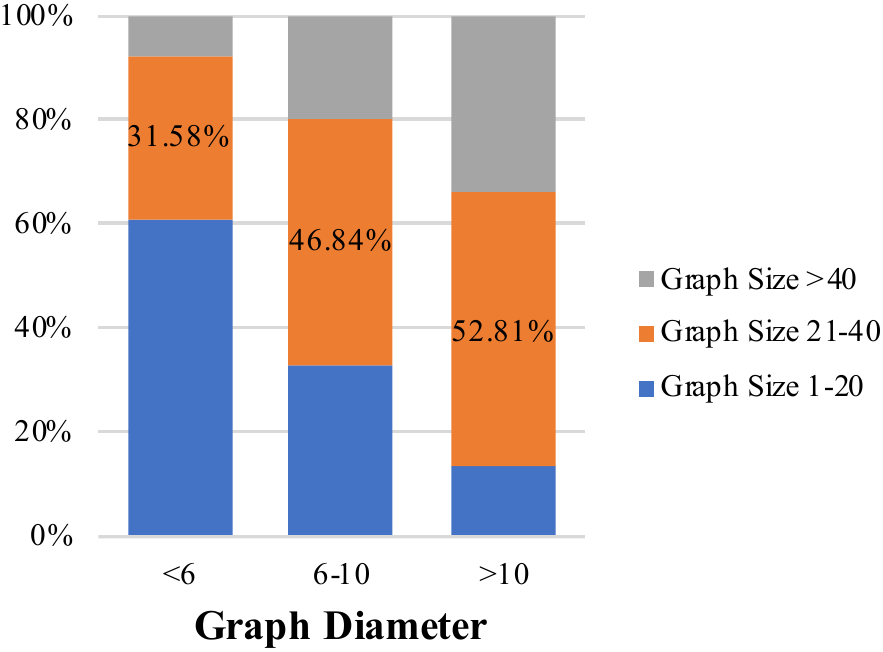}
    \end{minipage}
    
    \label{fig:diameter_on_graph_size}
    }

    \subfigure[The distribution of graph reentrancies by graph size.]{
    \begin{minipage}[t]{0.4\linewidth}
    \centering
    \includegraphics[width=1\columnwidth]{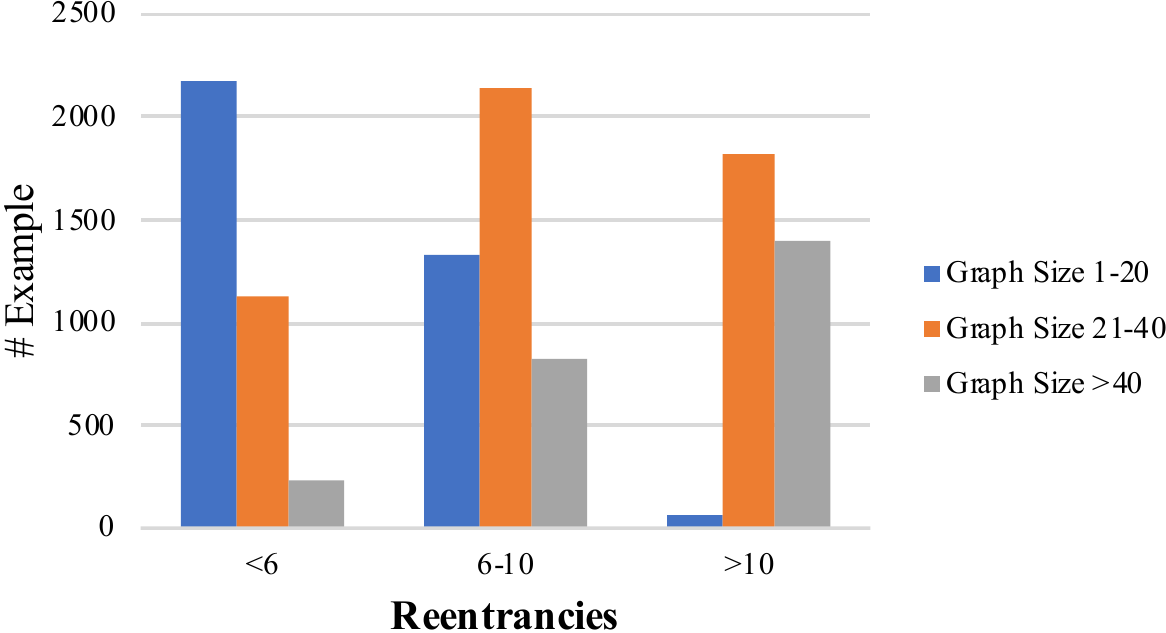}
    \end{minipage}
    
    \begin{minipage}[t]{0.3\linewidth}
    \centering
    \includegraphics[width=1\columnwidth]{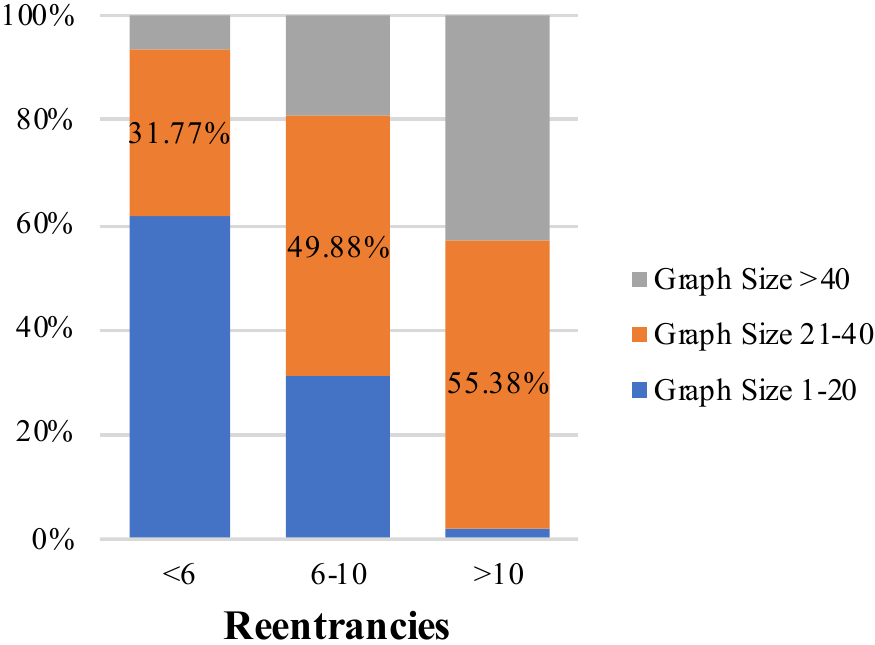}
    \end{minipage}
    
    \label{fig:reentrancies_on_graph_size}
    }

    \caption{The clustered column charts of graph diameter and  reentrancies by graph size.}
    \label{fig:distribution_on_graph_size}
\end{figure*}

\subsection{Distribution on Graph Size}
\label{append:distribution_on_graph_size}
On the ENT-DESC test set, previous study \cite{cheng-etal-2020-ent} and our experimental results (in Table \ref{tab:graph_stru_ent_dest}) suggest that the model performs the best when the graph size lies in the range of $21-40$ and has a poorer performance when the number of triples is too small or too large.
It should be due to the fact that the input information is insufficient or very noisy. 
However, we find that the model performance increases as the graph diameter and reentrancies increase. 
For further investigation, we calculate the distribution of graph diameter and reentrancies broken down by graph size, respectively. The results are summarized in Figure \ref{fig:distribution_on_graph_size}. As shown in Figure \ref{fig:diameter_on_graph_size}, the proportion of graphs with size $21-40$ increases as the graph diameter increases. As shown in Figure \ref{fig:reentrancies_on_graph_size}, the results on graph reentrancy follow a pattern similar to graph diameter. In a word, in ENT-DESC, the noise decreases as the graph diameter and reentrancies increase, so the model performs better.

\begin{figure}[t]
\centering
\includegraphics[width=1\columnwidth]{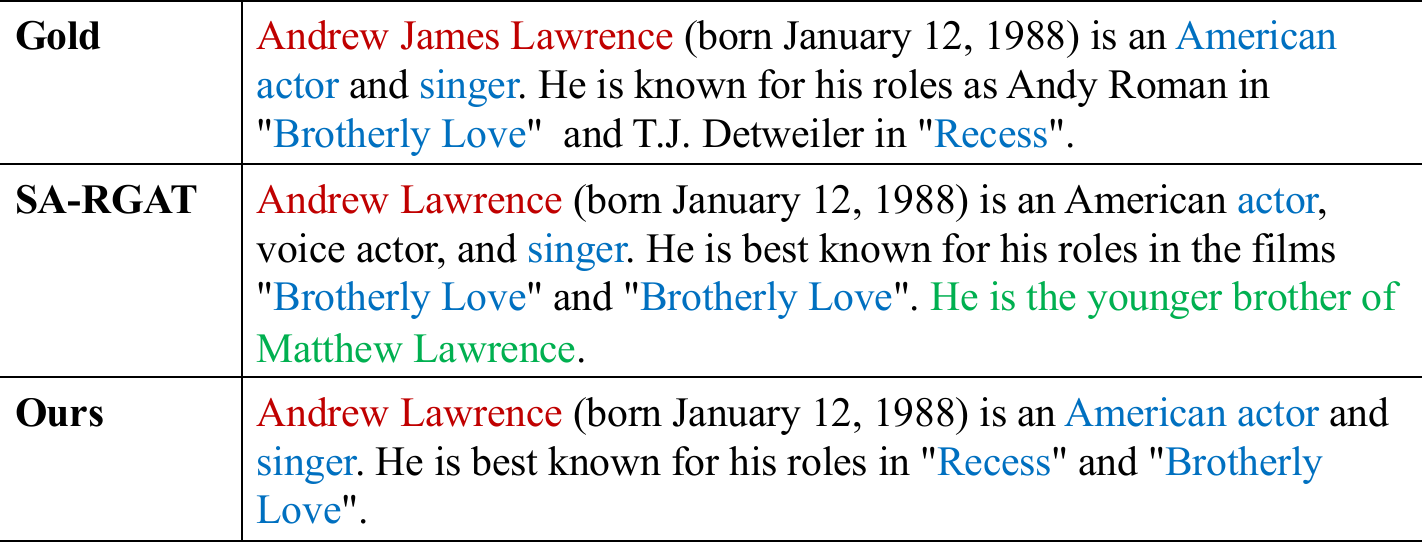}
\caption{An example of generated sentences. The main entity is highlighted in red, topic-related entities are highlighted in blue, and the sentence that is not faithful to the input graph is in green. }
\label{fig:case_study}
\end{figure}

\subsection{Case Study}
\label{append:case_study}
As shown in Figure \ref{fig:case_study}, we further take a typical example from our human study to better understand how our method improves the mode's performance. Given the Knowledge Graph containing the main entity ``Andrew Lawrence" and all its related entities, we aim to generate a description about the main entity. We notice that both the baseline and our model can identify the main entity. However, the baseline outputs a sentence describing the relation between ``Andrew Lawrence" and ``Matthew Lawrence". The relation is not existing in the input graph. Moreover, it repeatedly generates the entity ``Brotherly Love" and misses the related entity ``Recess". Compared with it, our model generates the sentences faithful to the input graph and correctly covers the main entity and most topic-related entities. We consider this is because the SACA helps the decoder obtain a more accurate input graph context, and the DGP removes the redundant nodes as the decoding stage progresses.

\subsection{Baseline Models}
\label{baseline}
On the AMR-to-Text task LDC2020T02, we compare our method with several baselines including:
\begin{itemize}
    \item \textbf{LDGCN} \cite{zhang2020lightweight} is a a dynamic fusion mechanism, which captures richer non-local interactions by synthesizing higher order information from the input graphs. A weight tied convolutions to reduce memory usage is applied.
    \item \textbf{SPRING} \cite{DBLP:conf/aaai/BevilacquaBN21} casts Text-to-AMR and AMR-to-Text as a symmetric transduction task and proposes a graph linearization and extending a pretrained encoder-decoder model.
\end{itemize}

On the KG-to-Text task ENT-DESC, we compare our method with several baselines including:
\begin{itemize}
    \item \textbf{s2s} \cite{bahdanau2015neural} is a encoder-decoder based model, which allows a model to automatically (soft-)search for parts of a source sentence that are relevant to predicting a target word, without having to form these parts as a hard segment explicitly.
    \item \textbf{GraphTransformer} \cite{koncel2019text} introduces a novel graph transforming encoder which can leverage the relational structure of such knowledge graphs without imposing linearization or hierarchical constraints.
    \item \textbf{GRN} \cite{beck2018graph} couples the recently proposed Gated Graph Neural Networks with an input transformation that allows nodes and edges to have their own hidden representations.
    \item \textbf{GCN} \cite{marcheggiani2018deep} proposes an alternative encoder based on graph convolutional networks that directly exploits the input structure. 
    \item \textbf{DeepGCN} \cite{guo2019densely} introduces a dense connection strategy, which is able to integrate both local and non-local features to learn a better structural representation of a graph.
    \item \textbf{MGCN + CNN} \cite{cheng-etal-2020-ent} is a multi-graph structure that is able to represent the original graph information more comprehensively. We do not report the results of MGCN + CNN + delex. Because it applies the delexicalization technique on the ENT-DESC dataset, which delexicalizes the main entity and topic-related entities by replacing these entities with tokens indicating the entity types and indices. The delexicalization technique greatly boosts their performance on ROUGE-L. They do not release the code about delexicalization, and we can not reproduce it. 
\end{itemize}
What's more, FINETUNE, ADAPT and SA-RGCN are T5-based models proposed in \citep{ribeiro-etal-2021-structural}. 

\end{document}